\documentclass[11pt]{article}

\usepackage[final]{acl}

\usepackage{times}
\usepackage{latexsym}

\usepackage[T1]{fontenc}

\usepackage[utf8]{inputenc}

\usepackage{microtype}

\usepackage{inconsolata}

\usepackage{graphicx}

\usepackage{enumitem}
\usepackage{amsmath}
\usepackage{amssymb}
\usepackage[T1]{fontenc}
\usepackage{textcomp}
\usepackage{lmodern}
\usepackage{booktabs}
\usepackage{pgfplots}
\usepackage{pgfplotstable}
\usepackage{subcaption}
\usepackage[most]{tcolorbox}
\usepackage{xcolor}
\usepackage{listings}

\definecolor{LMUGreen}{HTML}{00883A}
\definecolor{darkgreen}{rgb}{0.0, 0.5, 0.0}
\definecolor{coolblue}{rgb}{0.2, 0.4, 0.8}
\definecolor{coolorange}{rgb}{0.9, 0.4, 0.1}
\definecolor{coolred}{rgb}{0.8, 0.2, 0.2}

\title{Reinforcement Learning for Latent-Space Thinking in LLMs}

\author{
 \textbf{Enes Özeren\textsuperscript{1}},
 \textbf{Matthias Assenmacher\textsuperscript{1,2}}
\\
\\
 \textsuperscript{1}LMU Munich, Department of Statistics, Germany \\
 \textsuperscript{2}Munich Center for Machine Learning (MCML), Germany
\\
 \small{
   \textbf{Correspondence:} \href{mailto:enes.oezeren@campus.lmu.de}{enes.oezeren@campus.lmu.de}
 }
}

\begin{document}
\maketitle
\begin{abstract}
Chain-of-Thought (CoT) reasoning typically utilizes the discrete language space for thinking, which is inherently inefficient, as many generated tokens only enforce linguistic rules that are not required for reasoning. To bypass this, \textit{latent-space thinking} allows models to think using the continuous embedding space. While existing methods for training those models show domain-specific gains, they fail to maintain performance in complex tasks, such as mathematical reasoning. We experimentally demonstrate that the Coconut approach, a form of supervised fine-tuning for latent-space thinking, is highly sensitive to design choices and exhibits several inherent limitations. To address these issues, we investigate reinforcement learning (RL) techniques — an underexplored direction in latent-space thinking — including GRPO and design a novel \textit{Latent RL} method for directly optimizing the latent thinking steps. Our experimental results reveal that these RL-trained models still lag behind traditional language-space CoT models in the mathematical reasoning domain. We make our codebase publicly available.\footnote{\href{https://github.com/enesozeren/latent-space-thinking-model}{GitHub repository}}.
\end{abstract}

\section{Introduction}
\label{sec:intro}

\begin{figure}[ht]
    \centering
    \includegraphics[width=0.5\textwidth]{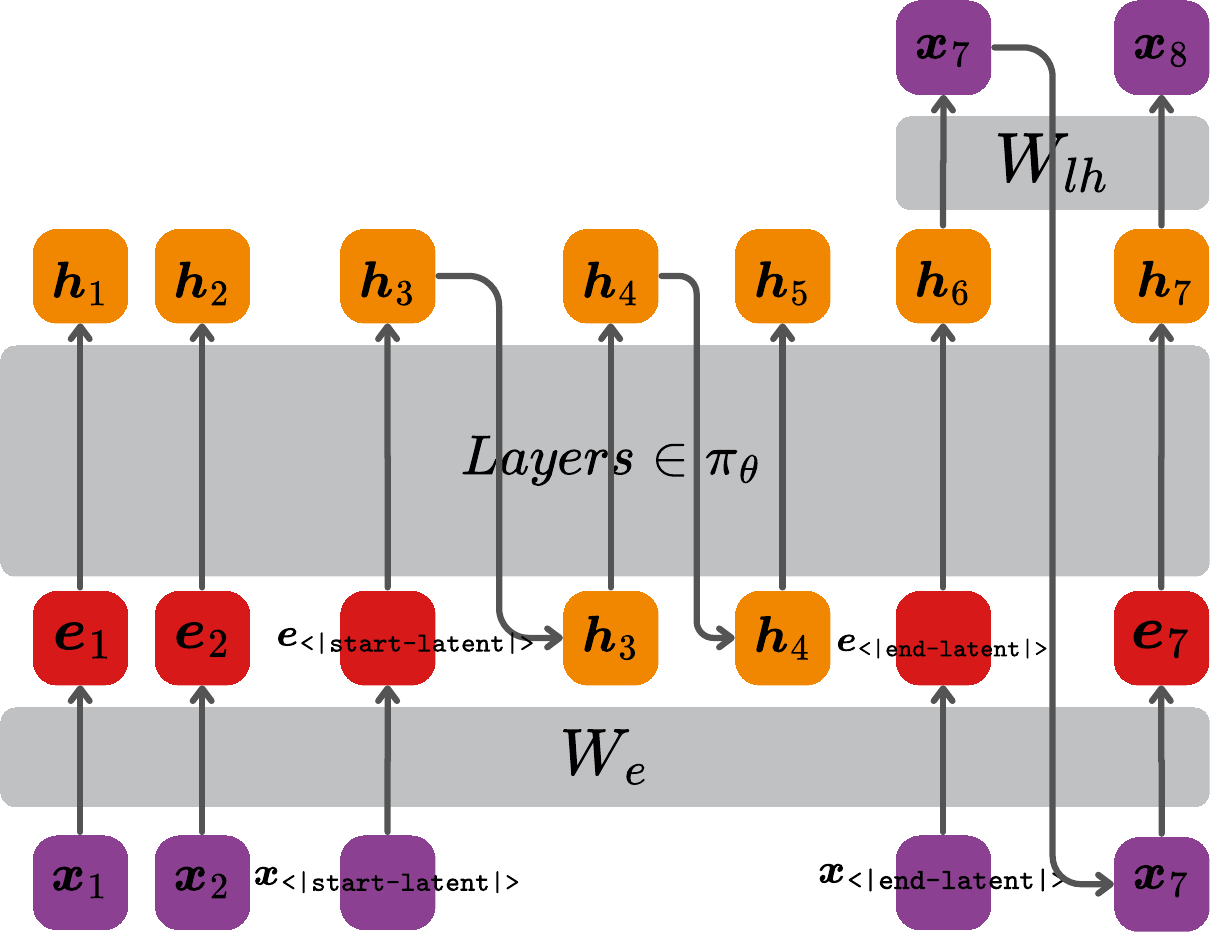}
    \caption{Illustration of latent-space thinking. Starting from the input ($\boldsymbol{x_1}$, $\boldsymbol{x_2}$), the \texttt{<|start-latent|>} is inserted, then the model switches to the latent mode and performs two latent steps ($C=2$). Once the \texttt{<|end-latent|>} token is reached, the model switches back to language mode to answer in language-space ($\boldsymbol{x_7}$, $\boldsymbol{x_8}$).}
    \label{fig:latent_space_thinking}
\end{figure}

Language model (LM) training is typically divided into two phases: pre-training, where models acquire general language understanding and world knowledge, and post-training, which adapts them for downstream use cases such as dialogue, code completion, or translation \citep{grattafiori2024llama}. Among the most widely used post-training methods is reinforcement learning (RL), which aligns LM behavior with human preferences \citep{ouyang2022training, touvron2023llama_b}. RL has also been applied to strengthen reasoning abilities in domains such as mathematics and programming \citep{shao2024deepseekmath}. While effective, such methods incur substantial computational costs, as models trained in this way tend to produce longer responses during inference \citep{guo2025deepseek, zheng2025group}.

An alternative direction improves efficiency by shifting reasoning from the language space (discrete tokens) to the latent space \citep[i.e., the continuous hidden states, see Fig. \ref{fig:latent_space_thinking},][]{hao2025training, zhu2025survey}, inspired by neuroscience findings that language primarily serves communication rather than thought \citep{fedorenko2024language}. A notable example is the \textit{Coconut} (Chain of Continuous Thought) method \citep{hao2025training}, which directly reuses final-layer embeddings as subsequent input, rather than decoding them into tokens. This allows the model to operate in its continuous latent space, avoiding the bottleneck imposed by discrete language tokens. However, training such models is non-trivial, as no ground-truth latent-space thinking\footnote{With "Latent-Space Thinking" we refer to the same concept coined as "Latent Reasoning" by \cite{hao2025training}. In our opinion, this is more adequate, as only a few latent steps are probably not yet real reasoning.} data exists. To address this, Coconut follows the strategy of \citet{deng2024explicit}, employing a supervised fine-tuning (SFT) approach where language-space reasoning steps are gradually replaced by latent-space thinking steps (referred to as \textit{Coconut SFT} procedure). A crucial aspect of this procedure is that latent thinking steps do not receive direct supervision signals; instead, they are guided only through indirect signals during training. As a result, even though \citet{hao2025training} reported improvements in the logical reasoning domain, the performance drops in the mathematics domain when compared with the language-space thinking models.

In this work, we explore RL-based methods as a way to improve latent-space thinking models. Initial experiments applying group relative policy optimization \citep[GRPO,][]{shao2024deepseekmath} show no measurable performance gains. As an alternative, we explore a novel RL method tailored to latent-space thinking to try to overcome the indirect learning signal limitation of the Coconut SFT method. However, this approach results in unstable training and fails to deliver performance improvements. Finally, we conduct an in-depth analysis of the Coconut SFT procedure, including ablation studies, which demonstrate that the procedure is sensitive to several factors, with performance being highly fragile. This analysis also evaluates several modifications, with the results indicating some promising adaptations as well as methodological weaknesses to be improved in future work.

The main contributions of this work are summarized as follows:

\begin{enumerate}
    \item An empirical study of training latent-space thinking models with GRPO.
    \item The introduction of a novel RL method tailored to latent-space thinking models.  
    \item An empirical in-depth analysis of the Coconut SFT procedure.    
\end{enumerate}

\section{Related Work}
\label{sec:related}

\paragraph{Latent-Space Thinking}

In their recent survey paper, \citet{zhu2025survey} highlight the growing body of work on latent-space thinking. One of the earliest studies by \citet{goyal2023think} demonstrated that inserting dummy \texttt{\textless pause\textgreater} \space tokens before generating a response improves LM performance on some tasks. This finding suggests that using more compute, even without producing language tokens, is beneficial for enabling the model to create richer (internal) representations before generating the answer. Similarly, \citet{pfau2024let} showed that even adding meaningless tokens like \texttt{‘.’} (dot) can achieve performance comparable to Chain-of-Thought (CoT) reasoning in certain problems.

Instead of focusing on latent thinking, \citet{deng2024explicit} aimed to internalize the CoT process within LMs. They proposed an SFT procedure that gradually removes explicit CoT steps from the training data, allowing the model to internalize the CoT process. Their results showed that this approach outperforms models trained without CoT, but still falls short of the performance achieved by models trained with CoT steps. Building on this idea, \citet{hao2025training} introduced the Coconut method, which incorporates latent thinking steps while also gradually removing the CoT steps during SFT. The goal is to enable the model to reason in an unrestricted latent space, rather than in the discrete language space. They define \textit{latent reasoning} as feeding the last hidden layer's last embedding back to the model as input, rather than decoding it into a token and feeding the corresponding token embedding in the next autoregressive iteration, as typically done in standard text generation. Their experiments show that this approach provides slight improvements on certain logical reasoning benchmarks but performs worse than explicit CoT in language space on the mathematical benchmark GSM8K \citep{cobbe2021training}. \citet{zhu2025reasoning} build on Coconut by developing a theoretical framework to explain why latent-space thinking can enhance LM performance on certain tasks, with a particular focus on the graph reachability problem.

\paragraph{Reinforcement Learning}

One of the first applications of RL for LM post-training was proximal policy optimization \citep[PPO,][]{schulman2017proximal}, which laid the groundwork for reinforcement learning from human feedback \citep[RLHF,][]{ziegler2019fine,ouyang2022training} for LMs. It aligns models with human preferences by increasing the likelihood of responses favored by annotators. Because collecting human feedback for every output is costly, a reward model is trained on limited annotations to assign a scalar reward to each complete response \citep{ouyang2022training}. However, this single score gives sparse feedback for long outputs. To refine this, a value model \citep{mnih2016asynchronous} predicts token-level contributions to the reward, offering more fine-grained feedback and improving training efficiency.

Recently, exploration of RL methods to enhance the reasoning capabilities of LMs has gained momentum \citep{shao2024deepseekmath, lambert2024t, guo2025deepseek, zheng2025group}. One of the first effective open-source approaches, \citet{shao2024deepseekmath}, introduced the GRPO method, which uses rule-based rewards tailored to mathematics and coding tasks—based on answer correctness or unit-test success. They reported consistent improvements during training and found that GRPO encourages progressively longer reasoning chains. Despite extensive exploration of SFT strategies in latent-space thinking models, the application of RL in this context remains largely unexplored.

\section{Methodology}
\label{sec:method}

\subsection{Latent Space Thinking}
\label{sec:methodology-latent-space-thinking}

A LM $\pi_\theta$ with parameters $\theta$ processes text by tokenizing it into $t$ tokens, each represented as a one-hot vector over the vocabulary $V$, forming $X^t \in \mathbb{R}^{t \times |V|}$. Tokens are embedded via the embedding matrix $W_e \in \mathbb{R}^{|V| \times D}$:
\begin{equation}
    E^t = X^t W_e
\end{equation}

The embeddings are contextualized by the model layers:
\begin{equation}
    H^t = Layers(E^t), \quad H^t \in \mathbb{R}^{t \times D}
\end{equation}

To predict the next token, the final hidden state $\boldsymbol{h}_t$ is projected into the vocabulary space using the language head $W_{lh} \in \mathbb{R}^{|V| \times D}$:
\begin{equation}
    \boldsymbol{y}_t = W_{lh}\boldsymbol{h}_t, \quad \boldsymbol{p}_t = \mathsf{softmax}(\boldsymbol{y}_t)
\end{equation}

A decoding strategy selects the next token, and the process repeats auto-regressively until the end-of-sequence token is generated.

Latent-space thinking, as proposed by \citet{hao2025training}, modifies the first and final step of the autoregressive generation process as illustrated in Figure~\ref{fig:latent_space_thinking}. As a starting point, three new tokens, \texttt{<|start-latent|>}, \texttt{<|latent|>}, \texttt{<|end-latent|>}, are added to the vocabulary and their embeddings are initialized randomly in the embedding matrix $W_e$. The \texttt{<|latent|>} acts purely as a placeholder for latent steps, and its embedding is never used by the model. When an input text $X^t$ is provided, the \texttt{<|start-latent|>} token is directly appended as the $(t+1)$-th token, forming $X^{t+1}$.

Instead of proceeding with token generation using the language head matrix $W_{lh}$ and a decoding strategy, the latent-space thinking model performs a latent step by taking the last contextualized token embedding $\boldsymbol{h}_{t+1}$ and appending it directly to the embedded sequence $E^{t+1}$. This is done via row-wise concatenation operation, denoted as $\text{concat}$:
\begin{equation}
    E^{t+2} = \text{concat} (E^{t+1},  \boldsymbol{h}_{t+1})
\end{equation}

This process, referred to as latent-space thinking, is repeated until a predefined number $C$ of latent steps have been performed. After that, the \texttt{<|end-latent|>} token is appended. Following this, the latent-space thinking model reverts to the standard token generation in language space. To implement this mechanism, we used the Huggingface transformers library \citep{wolf2019huggingface} with targeted modifications to the \texttt{generate} and \texttt{forward} methods.

\subsection{Existing Training Methods}
\label{sec:train_exist}

\paragraph{Supervised Fine-Tuning} For the language-space thinking model, we apply standard SFT without any modifications. Given a question-response pair $(X^{q_i}, X^{r_i})$, we do one forward pass and calculate the cross-entropy loss on each response token:
\begin{equation}
    \mathcal{L}_{\mathrm{SFT}} = - \sum_{i=1}^N \log \left( \pi_\theta(X^{r_i} \mid X^{q_i}) \right)
\end{equation}

For the latent-space thinking model, we follow the Coconut SFT procedure \citep{hao2025training}. Given $(X^{q_i}, X^{r_i})$, the model first processes all tokens in the question $X^{q_i}$, then generates latent steps auto-regressively as described in Section~\ref{sec:methodology-latent-space-thinking}. After the latent steps are generated, the embeddings of the question tokens $E^{q_i}$, the latent steps $E^{l_i}$, and the response tokens $E^{r_i}$ are concatenated and passed through the model in a single forward pass. During training, a cross-entropy loss is computed over the response tokens only; question tokens and latent steps are masked and excluded from the loss. Thus, the objective minimized during the Coconut SFT procedure is:
\begin{equation}
    \mathcal{L}_{\mathrm{SFT}} = - \sum_{i=1}^N \log \left( \pi_\theta(X^{r_i} \mid X^{q_i}, E^{l_i}) \right)
\end{equation}

This method optimizes the latent steps indirectly through the language space tokens in the response part. 

\paragraph{(Modified) GRPO} Applying RL to latent-space thinking models represents an unexplored research direction. We therefore experiment with the GRPO algorithm to investigate its impact. The language-space thinking model is trained with the GRPO implementation from the Hugging Face TRL package \citep{vonwerra2022trl}. In contrast, training the latent-space thinking model with GRPO requires a modified version of this module. We modify the TRL package to exclude latent-step embeddings from the GRPO loss calculations since the GRPO loss requires token probabilities.

Similar to the Coconut SFT procedure, this modified GRPO setup optimizes the latent steps indirectly through the probabilities of tokens in the language space, and thus only through the final answer portion. In GRPO training, two types of rewards are used: an accuracy and a format reward (see Appendix~\ref{sec:appendix-reward-funcs}).

\subsection{Proposing Latent RL}
\label{sec:last-latentrl}

Neither Coconut SFT nor GRPO provides explicit supervision for training latent steps, as it is not straightforward to determine what constitutes a “good” latent step, since these steps are represented by continuous vectors and no ground-truth latent vectors are available.

To address this challenge, we propose an alternative RL approach, referred to as \textit{Latent RL}. In the Latent RL training setup, a value model $\rho_{\theta+\phi}$ is employed to directly optimize the latent steps. The value model $\rho_{\theta+\phi}$ shares some of its parameters $\theta$ with the underlying policy model $\pi_\theta$ and is extended with additional value head parameters $\phi$ on top.

\begin{figure}[ht]
    \centering
    \includegraphics[width=0.48\textwidth]{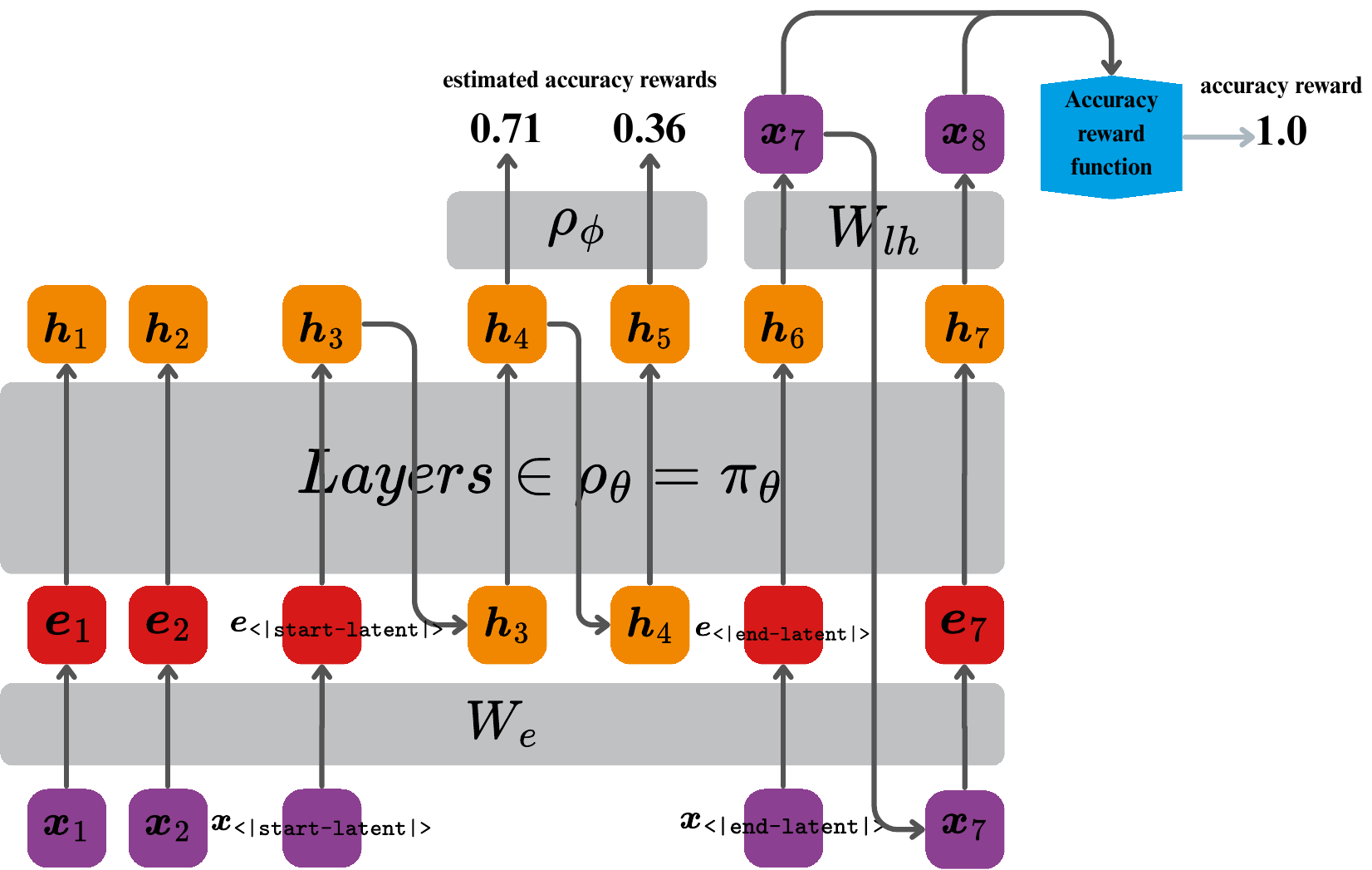}
    \caption{Illustration of Latent RL training. The value model head estimates the accuracy reward for each latent step.}
    \label{fig:latent_rl}
\end{figure}

The Latent RL training procedure is illustrated in Figure~\ref{fig:latent_rl}. In the first phase, we train $\rho_{\theta+\phi}$ by freezing parameters $\theta$. In the second phase, we train the policy model by utilizing the value model. To achieve this, we generate responses with latent steps for given questions and then maximize the value model predictions:
\begin{equation}
    \arg\min\limits_{\theta} L_{LAT}, \quad  L_{LAT} = -\rho_{\theta+\phi}(\boldsymbol{h})
    \label{eq:latent-rl-objective}
\end{equation}

Further details of the Latent RL procedure can be found in Appendix~\ref{sec:appendix-latent-rl-methodology}.

\subsection{Performance Measures}

We primarily evaluate LMs for their mathematical capabilities and efficiency. Hence, the following two evaluation metrics are employed.

\paragraph{pass@k} This metric estimates the expected probability that a model generates a correct answer to a question within its first $k$ attempts \citep{chen2021evaluating, guo2025deepseek, dang2025reinforcement}. It is defined as
\begin{equation}
    \mathrm{pass@}k
    =
    \mathbb{E}_{\text{problems}}
    \Biggl[
    1 \;-\;\frac{\binom{n - c}{k}}{\binom{n}{k}}
    \Biggr],
    \label{metric:pass@k}
\end{equation}

where $n$ denotes the number of generated answers for a given question and $c$ denotes the number of correct answers among them. We set $n = 4$ and report both pass@1 and pass@4.    
    
\paragraph{\# Tokens} The average number of tokens generated by the LM to produce an answer is a measure of output length and efficiency.

\section{Experimental Setup}

\subsection{Models}

The base LM used in this study is Qwen2.5 1.5B Base \citep{yang2024qwen2}. This choice strikes a balance between computational feasibility and model capability in this study. For brevity, the model is referred to simply as Qwen Base. All methods and implementations proposed here are model-agnostic and can be applied to other decoder-only transformer LMs.

\begin{table}[ht]
    \centering
    \small
    \begin{tabular}{ll}
    \toprule
    \textbf{Model Name} & \textbf{Training} \\
    \midrule
    \small{Qwen Base} & \tiny{Base model} \\
    \midrule \midrule
    \small{Qwen SFT} & \tiny{Qwen Base + SFT} \\
    \small{Qwen SFT \& GRPO} & \tiny{Qwen SFT + GRPO } \\
    \midrule \midrule
    \small{LaST SFT} & \tiny{Qwen Base + Coconut SFT} \\
    \small{LaST SFT \& GRPO} & \tiny{LaST SFT + GRPO} \\
    \small{LaST SFT \& Latent RL} & \tiny{LaST SFT + Latent RL} \\
    \bottomrule
    \end{tabular}
    \caption{Models compared in this study. LaST is short for \textit{\textbf{La}tent-\textbf{S}pace \textbf{T}hinking Model}.}
    \label{tab:models}
\end{table}

To analyze the effects of different training methodologies, we train and compare the models as outlined in Table~\ref{tab:models}.

\subsection{Datasets}

Since we employ two SFT and RL training methodologies, each with distinct requirements for the training data, we utilize different datasets for each of them (see Table~\ref{tab:training_testing_dataset_volumes}).

\begin{table}[ht]
    \centering
    \small
    \begin{tabular}{ll|c}
    \toprule
    \textbf{Dataset} & \textbf{Purpose} & \textbf{Usage Vol.} \\
    \midrule
    GSM8K-Aug-NL     & Training (SFT) & 300K \\
    OpenR1-Math-220k & Training (RL) & 10K  \\
    GSM8K            & Testing       & 1.3K \\
    MATH500          & Testing       & 500  \\
    \bottomrule
    \end{tabular}
    \caption{The volumes of the datasets used in this study. K denotes thousands. For the training part, full datasets are not utilized for efficiency purposes. The Usage Vol. column reports the amount we used in our experiments.}
    \label{tab:training_testing_dataset_volumes}
\end{table}

For SFT, the primary requirement is that the dataset must contain question-response pairs. The Coconut SFT procedure, however, has the additional requirement that the responses $X^{r_i}$ must be structured as a list of sequential steps. This means that, e.g., the response should contain a sequence of sentences or mathematical formulations, each corresponding to an individual, \textit{separate} reasoning step. To satisfy these requirements, the GSM8K-Aug-NL dataset \citep{deng2023implicit} is selected for SFT.

For pure RL training methods, a dataset containing question-answer pairs $(X^{q_i}, X^{a_i})$ is sufficient; the full response\footnote{We strictly differentiate \textit{answers} (just the final result) from \textit{responses} (including the whole reasoning trace) to emphasize the different requirements.} is not necessary. We use OpenR1-Math-220k \citep{openr1}.

\subsection{Training Details}
\label{sec:experimental-setup-training-details}

The models in Table~\ref{tab:models} are trained using the datasets in Table~\ref{tab:training_testing_dataset_volumes}. The details of the training are given in this section. All the training hyperparameters are provided in Appendix~\ref{sec:appendix-hps}.

\paragraph{Qwen SFT.} This model is obtained by SFT training of the Qwen Base model on 300K question-response pairs from GSM8K-Aug-NL.

\paragraph{Qwen SFT \& GRPO.} This model is trained with the GRPO algorithm using rule-based rewards, starting from the Qwen SFT model. The training is performed on 10K samples from OpenR1-Math-220k.

\paragraph{LaST SFT.} This model is trained using the Coconut SFT procedure proposed by \citet{hao2025training}, starting from the Qwen SFT model and using the same 300K question-response pairs from GSM8K-Aug-NL as in the Qwen SFT training process. This procedure follows a multi-stage approach: in each stage, one language step (typically one sentence) from the ground-truth response is removed and replaced with two latent steps. In the final stage, all language steps are removed, and the model relies solely on latent steps to predict the final answer. Consistent with \citet{hao2025training}, a three-stage approach was adopted, allowing the model to use up to six latent steps.

A key difference to \citet{hao2025training} lies in the choice of training data: instead of the GSM8K-Aug dataset, which contains only mathematical calculation steps, the GSM8K-Aug-NL dataset was used, where the steps consist of both language tokens and mathematical calculations. This choice was made to align with the main motivation for introducing latent steps - namely, to enable the model to internalize not only mathematical calculations but also language.

\paragraph{LaST SFT \& GRPO.} This model is trained on the same 10K question-answer pairs from the OpenR1-Math-220k dataset as used for the Qwen SFT \& GRPO model, but starting from the LaST SFT model.

\begin{table*}[ht]
    \setlength{\tabcolsep}{2pt}
    \centering
    \begin{tabular}{cc|ccc|ccc}
        \toprule
        \textbf{Model} & \textbf{Shot} &
        \multicolumn{3}{c|}{\textbf{GSM8K}} & 
        \multicolumn{3}{c}{\textbf{MATH500}} \\
        & & \textbf{pass@1} & \textbf{pass@4} & \textbf{\# Tokens} & \textbf{pass@1} & \textbf{pass@4} & \textbf{\# Tokens} \\
        \midrule
        Qwen Base & Zero & 24.8 & 49.7 & 156 & 12.1 & 27.4 & 310 \\
        Qwen Base & One & \textbf{57.0} & \textbf{73.4} & 575 & \textbf{25.4} & \textbf{40.2} & 589 \\
        \midrule \midrule
        Qwen SFT & Zero & 62.6 & 76.0 & 105 & 22.0 & 32.0 & 290 \\
        Qwen SFT \& GRPO & Zero & \textbf{72.6} & \textbf{82.5} & 216 & \textbf{38.7} & \textbf{50.0} & 366 \\
        \midrule \midrule
        LaST SFT & Zero & 22.6 & 29.9 & 19 & 8.5 & 12.4 & 19 \\
        LaST SFT \& GRPO & Zero & 21.8 & 24.1 & 19 & \textbf{9.2} & \textbf{14.0} & 18 \\
        LaST SFT \& Latent RL & Zero & \textbf{22.7} & \textbf{30.1} & 19 & 8.4 & 13.2 & 19 \\
        \bottomrule        
    \end{tabular}
    \caption{Evaluation results on GSM8K and MATH500 benchmarks.}
    \label{tab:overall_performance_results}
\end{table*}

\paragraph{LaST SFT \& Latent RL.} The Latent RL method consists of two stages. First, the value model is trained to estimate accuracy rewards for latent steps. For this purpose, 20K question-answer pairs from the OpenR1-Math-220k are selected, and responses are generated with the LaST SFT model. Accuracy rewards for the generated responses are computed using the reward function, and the value model is trained to predict these rewards from the latent steps. The value head itself consists of a linear layer that maps each latent step embedding to a scalar score representing the estimated reward. In the second stage, training is performed using 10K question-answer pairs from the OpenR1-Math-220k dataset, distinct from those used for value model training. These pairs are the same pairs as used in the LaST SFT \& GRPO model training to be consistent.

\section{Results}

This chapter summarizes the main conclusions drawn from the experiments (cf. Table ~\ref{tab:overall_performance_results}) and an ablation study for Coconut SFT.

\subsection{Main Results}

\paragraph{Evaluating Qwen Base is challenging.} In the zero-shot setting, the Qwen Base model often fails to follow the instructed answer format, making it difficult to evaluate performance based on boxed answers in the answer block (see Appendix~\ref{sec:appendix-instruction-following} for more details). To address this, the model was also tested in a one-shot setting with a single example showing the expected format. However, prior work shows that few-shot prompting, including one-shot, can boost performance beyond format compliance \citep{brown2020language, wei2022chain}, so the model’s true zero-shot ability likely lies between its zero- and one-shot results presented here. Notably, in one-shot evaluation, the model sometimes hallucinated additional questions and answers, resulting in notably longer responses.

\paragraph{SFT on GSM8K-Aug-NL improves the performance on GSM8K.}
The Qwen SFT model outperforms the Qwen Base model under both zero-shot and one-shot prompting on the GSM8K benchmark. This improvement can be attributed to the use of the GSM8K-Aug-NL dataset during SFT, which contains questions similar to those in GSM8K and thereby enhances domain-specific performance. On the MATH500 benchmark, however, the Qwen SFT model surpasses the Qwen Base model under zero-shot prompting but performs worse than under one-shot prompting. This indicates that the fine-tuning provides less benefit for MATH500, suggesting more limited transfer compared to the gains observed on GSM8K.

\paragraph{GRPO on OpenR1-Math-220k improves the performance and increases the response length.}
The Qwen SFT \& GRPO model, obtained through training with the GRPO algorithm with rule-based rewards, achieves higher performance on both benchmarks compared to the Qwen SFT model. In addition, the average response length increases after GRPO training, consistent with previous studies \citep{guo2025deepseek, dang2025reinforcement}.

\paragraph{Coconut SFT harms model performance.}
Similar to the findings in \citet{hao2025training}, the LaST SFT model trained with the Coconut SFT procedure performs worse than the Qwen SFT model on mathematical benchmarks. And since the model only produced six latent steps prior to the answer block, the number of tokens generated is significantly lower than other language-space thinking models.

\begin{figure}[ht]
\centering
\begin{tikzpicture}
    \begin{axis}[
        width=8cm,
        height=5.5cm,
        xlabel={Number of Latent Steps},
        ylabel={pass@1},
        xlabel style={yshift=0.3em},
        ylabel style={yshift=-1.0em},
        xmin=1, xmax=128,
        ymin=0, ymax=27,
        xmode=log,
        log basis x=2, 
        xtick={2, 4, 6, 16, 64},
        xticklabels={2, 4, 6, 16, 64}, 
        ytick={0, 10, 20, 30},
        grid=both,
        grid style={dashed,gray!30},
        line width=1pt,
        mark size=3pt,
        legend style={
            at={(0.02,0.02)},
            anchor=south west,
            legend columns=1,
            font=\tiny,
        },
        nodes near coords,
        every node near coord/.append style={
            font=\small, 
            yshift=2pt
        }
    ]
    
    \addplot[color=LMUGreen, mark=*, solid] coordinates {
        (2, 22.2) (4, 23.0) (6, 22.6) (16, 22.3) (64, 2.7)
    };
    \addlegendentry{GSM8K}
    
    \addplot[color=LMUGreen, mark=square*, dotted] coordinates {
        (2, 8.3) (4, 8.5) (6, 8.5) (16, 10.0) (64, 0.8)
    };
    \addlegendentry{MATH500}
    
    \end{axis}
\end{tikzpicture}
\caption{Effect of varying the number of latent steps on the pass@1 score with the LaST SFT model. The x-axis has the log scale.}
\label{fig:latent_step_impact_on_performance}
\end{figure}
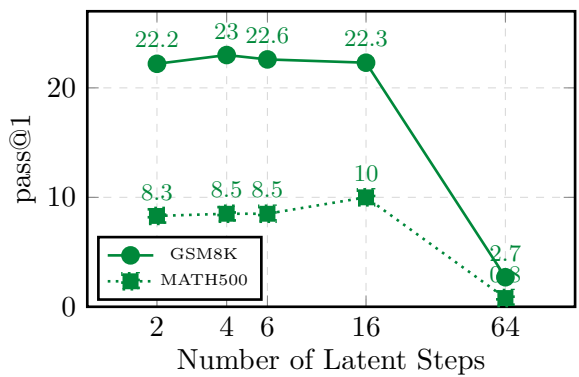

To examine the influence of the number of latent steps in the LaST SFT model, evaluations were conducted using configurations of 2, 4, 6, 16, and 64 latent steps, despite the model having been trained with a maximum of 6 latent steps (Figure~\ref{fig:latent_step_impact_on_performance}). The results show that increasing the number of latent steps does not improve performance. Although the model stays within distribution up to 16 latent steps, its performance declines sharply at 64 latent steps.

\paragraph{GRPO training does not improve the performance of LaST SFT.}
Using the GRPO algorithm, which ignores the latent steps, does not yield performance gains for the LaST model, in contrast to results for the Qwen SFT \& RL model. Two main factors may explain this outcome: 

First, the GRPO algorithm is designed for language-space reasoning models, where it increases the probability of generating intermediate language tokens that eventually lead to correct answers. In the LaST SFT model, however, no such intermediate language tokens are produced before the final answer, which prevents the learning signal from being effective. To address this limitation, we introduced Latent RL in Section~\ref{sec:last-latentrl}. Second, GRPO typically enhances performance by encouraging models to generate longer responses before the final answer, thereby increasing inference-time computation. In the LaST SFT model, the number of latent steps is fixed. However, as shown in Figure~\ref{fig:latent_step_impact_on_performance}, simply increasing the number of latent steps does not enhance performance. This indicates that the effect of this factor is limited.

\paragraph{Latent RL training is unstable.} 
Even though we address the limitations of other training techniques for latent-space thinking model training, we observe that Latent RL training shows unstable optimization dynamics (Appendix~\ref{sec:appendix-latent-rl-training-curves}). This is also reflected in the benchmark evaluations, where performance does not increase despite the Latent RL training. These observations suggest that the unstable Latent RL struggles to improve latent-space thinking. Exemplary model outputs for a question are presented in Appendix~\ref{sec:appendix-ex-responses}.

\subsection{Coconut SFT Ablations}

We conduct ablation studies for the Coconut SFT procedure, as the baseline latent-space thinking model is trained with it.

\paragraph{Necessity of the Coconut SFT Procedure.}
When latent-space thinking is introduced to the Qwen Base model without any additional training, the model produces out-of-distribution and nonsensical outputs (Figure~\ref{fig:model_ex_responses_nonsense}). This occurs because the model is fed its own latent steps as input—data it has never encountered during pre-training. Therefore, an SFT procedure is essential before applying any RL training, as RL relies on having at least some correct responses. To this end, we employ the Coconut SFT method \citep{hao2025training}.

\begin{figure}[htbp]
    \centering
    \begin{minipage}{0.49\textwidth}
        \begin{tcolorbox}[
            colback=gray!5,
            colframe=gray!30,
            boxrule=0.8pt,
            arc=3pt,
            left=8pt,
            right=8pt,
            top=1pt,
            bottom=1pt,
            title={\textbf{Qwen Base with Latent-Steps}},
            fonttitle=\small\sffamily,
            coltitle=gray!70!black
        ]
        \begin{lstlisting}[
            breaklines=true,
            breakindent=0pt,
            aboveskip=0pt, belowskip=0pt,
            breakatwhitespace=true,
            basicstyle=\scriptsize\ttfamily,
            frame=none,
            xleftmargin=0pt,
            xrightmargin=0pt
        ]
<|start-latent|><|latent|><|end-latent|> [\{\})\$);,;)\{degrees\}  ( (); shi zi equ zi . hua\})<|end-latent|><|end-latent|>) .all\_()\{\}
        \end{lstlisting}
        \end{tcolorbox}
    \end{minipage}
    \caption{Qwen Base response to the question in Figure~\ref{fig:model_ex_responses_question} after introducing latent-steps.}
    \label{fig:model_ex_responses_nonsense}
\end{figure}

\paragraph{Language Removal Approach.}
Instead of a step-wise removal of pre-defined reasoning steps (as in Coconut SFT), we investigate a token-wise removal approach, where a fixed number of tokens is removed. This relaxes dataset constraints, enabling the use of any question-response dataset for latent-space thinking model training beyond GSM8K-Aug-NL. For this investigation, all reasoning steps in GSM8K-Aug-NL answers are concatenated. During LaST SFT training, we remove 12 tokens per latent step—half the average step length—to match the step-wise setup.

\begin{table}[ht]
    \small
    \centering
    \begin{tabular}{c|c|c}
        \toprule
        \textbf{Removal Technique} & \shortstack{\textbf{GSM8K}\\\textbf{pass@1}} & \shortstack{\textbf{MATH500}\\\textbf{pass@1}} \\
        \midrule
        Step-wise & \textbf{17.9} & \textbf{11.3} \\
        \midrule
        Token-wise & 7.5 & 3.4
        \\        
        \bottomrule        
    \end{tabular}
    \caption{Investigation of different language token removal techniques in Coconut SFT.}
    \label{tab:language_removal_technique}
\end{table}

Step-wise removal performs better (Table~\ref{tab:language_removal_technique}), presumably because token-wise removal can truncate reasoning steps mid-thought, making it more difficult for the model to internalize the reasoning process.

\paragraph{Number of Epochs per Stage.} The Coconut SFT method uses 3 epochs per stage, with each stage involving one language step removal and two latent space insertions. We explore different numbers of epochs per stage:

\begin{table}[ht]
    \small
    \setlength{\tabcolsep}{4pt}
    \centering
    \begin{tabular}{c|c|c}
        \toprule
        \textbf{Epochs/Stage} & \shortstack{\textbf{GSM8K}\\\textbf{pass@1}} & \shortstack{\textbf{MATH500}\\\textbf{pass@1}} \\
        \midrule
        1 & \textbf{17.9} & \textbf{11.3} \\
        \midrule
        2 & 17.0 & 10.4
        \\        
        \bottomrule        
    \end{tabular}
    \caption{Ablation study on num. of epochs/stage.}
    \label{tab:numb_epochs_per_stage}
\end{table}

Training the LaST model with one epoch per stage performs better than with two epochs (Table~\ref{tab:numb_epochs_per_stage}). In comparison, the Coconut method \citep{hao2025training} reported using 3 epochs per stage. Further experiments were omitted due to the slow training of Coconut SFT.

\begin{figure}[ht]
\centering
\begin{tikzpicture}
    \begin{axis}[
        width=8cm,
        height=4.5cm,
        xlabel={Num. of Epochs in Last Stage},
        ylabel={pass@1},
        xlabel style={yshift=0.3em},
        ylabel style={yshift=-1.0em},        
        xmin=0, xmax=4,
        ymin=0, ymax=27,
        xtick={0,1,2,3,4},
        ytick={0,10,20,30},
        grid=both,
        grid style={dashed,gray!30},
        line width=1pt,
        mark size=3pt,
        legend style={
            at={(0.02,0.02)},
            anchor=south west,
            legend columns=1,
            font=\tiny,
        }
    ]
    \addplot[
        color=darkgreen, 
        mark=*,
        solid,
    ] coordinates {
        (1, 17.9) (2, 19.0) (3, 22.6)
    };
    \addlegendentry{GSM8K}
    \addplot[
        color=darkgreen, 
        mark=square*,
        dotted,
    ] coordinates {
        (1, 11.3) (2, 8.0) (3, 8.5)
    };
    \addlegendentry{MATH500}
    \end{axis}
\end{tikzpicture}
\caption{The effect of number of epochs in the last stage of the Coconut SFT procedure.}
\label{fig:number_of_epochs_for_last_stage}
\end{figure}
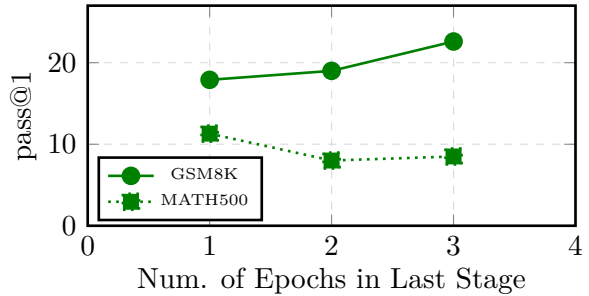

\paragraph{Final Stage Training Epochs Effect.} We also explore the impact of each epoch on the final stage of Coconut SFT procedure, showing that this last stage enhances the performance of the LaST SFT model on the GSM8K benchmark (Figure~\ref{fig:number_of_epochs_for_last_stage}). However, the observed performance drop on the MATH500 benchmark indicates that extended training on the GSM8K-Aug-NL dataset reduces the generalization.

\section{Conclusion}

In this work, we explored different training strategies for latent-space thinking models, starting from the Coconut SFT procedure \citep{hao2025training} and extending it to RL-based methods. While Coconut SFT provides a structured way to train latent-space thinking models, it consistently underperforms compared to language-space models on mathematical benchmarks. We showed, that its success depends heavily on design choices, such as the number of epochs in the final stage and token removal strategies. The method’s main limitation lies in the absence of direct learning signals to latent steps and its reliance on auto-regressive processing, which limits scalability.

RL approaches were then evaluated as potential improvements. Extending GRPO to Coconut-trained latent-space thinking models brought no gains, since GRPO also lacks direct latent-step supervision. To overcome this, we introduced the Latent RL method, which provides explicit learning signals to latent steps through a value model. However, training was unstable and did not improve benchmark performance.

Future work should aim to enhance SFT training efficiency, stabilize Latent RL, and develop mechanisms for adaptive control over latent steps. Overall, the findings suggest that current latent-space thinking training methods still lag behind language-space models, primarily due to their instability or the absence of supervision for latent-space thinking.

\section*{Limitations}

This study examines training techniques for latent-space thinking models in the mathematical domain. While this focus allows for precise evaluation, it limits generalizability to other tasks such as general knowledge, logical reasoning, or programming. Prior work \citep{hao2025training, zhu2025reasoning} suggests that latent-space thinking may perform differently in those domains, highlighting a key direction for future research.

Another limitation is the use of a single base model, which ensures consistency but narrows the scope of conclusions. Models with different sizes, architectures, or pretraining strategies may respond differently to these training methods, and exploring such variations could offer deeper insight.

A further challenge is efficiency: unlike standard LMs, latent-space thinking models require auto-regressive generation during both training and inference, increasing computational cost and limiting scalability. This makes large-scale training impractical with current setups.

Finally, latent-space thinking reduces model interpretability, making it harder to analyze internal processes or ensure transparency in deployment. Future work should therefore focus on improving interpretability—such as by studying attention or representation changes across latent steps—to better understand and control latent-space thinking behavior.

\section*{Acknowledgments}

We gratefully acknowledge the Leibniz Supercomputing Centre (LRZ) of the Bavarian Academy of Sciences and Humanities and the Munich Center for Machine Learning (MCML) for generously providing computational resources. Matthias Aßenmacher received funding from the Deutsche Forschungsgemeinschaft (DFG, German Research Foundation) under the National Research Data Infrastructure – NFDI 27/1 - 460037581 - BERD@NFDI.

\bibliography{custom}

\clearpage

\appendix

\section{Reward Functions}
\label{sec:appendix-reward-funcs}

For the GRPO method, accuracy and format rewards are used; while the accuracy reward is the same, the format reward differs as indicated below. The Latent RL method uses the same reward functions as GRPO for the latent-space thinking model.

\paragraph{Accuracy Reward}
\begin{center}
\small
\begin{tabular}{r@{\hspace{0.5em}}|l}
\toprule
\textbf{Reward} & \textbf{Condition} \\
\midrule
$r = 1$ & The correct answer appears \\ 
        & 
        in \texttt{\textbackslash boxed\{...\}} inside \\
        & \texttt{<answer>}\ldots\texttt{</answer>} tags. \\
\addlinespace
$r = 0$ & Otherwise. \\
\bottomrule
\end{tabular}
\end{center}
\vspace{0.5em}
The correctness is verified by exact string match or mathematical equivalence similar to \citet{openr1}.

\paragraph{Format Reward — Language-Space Thinking Models}
\begin{center}
\small
\begin{tabular}{r@{\hspace{0.1em}}|l}
\toprule
\textbf{Reward} & \textbf{Format} \\
\midrule
$r = 1$ & \texttt{<think>}\ldots\texttt{</think>} \\ & \texttt{<answer>} \texttt{\textbackslash boxed\{...\}} \texttt{</answer>} \\
\addlinespace
$r = 0.5$ & Same format but the \\ & \texttt{\textbackslash boxed\{...\}} is missing. \\
\addlinespace
$r = 0$ & Otherwise. \\
\bottomrule
\end{tabular}
\end{center}

\paragraph{Format Reward — Latent-Space Thinking Models}
\begin{center}
\small
\begin{tabular}{r@{\hspace{0.1em}}|l}
\toprule
\textbf{Reward} & \textbf{Format} \\
\midrule
$r = 1$ & \texttt{<|start-latent|>}\ldots\texttt{<|end-latent|>} \\ & \texttt{<answer>} \texttt{\textbackslash boxed\{...\}} \texttt{</answer>} \\
\addlinespace
$r = 0.5$ & Same format but the \\ & \texttt{\textbackslash boxed\{...\}} part is missing. \\
\addlinespace
$r = -1$ & Any of \texttt{<|start-latent|>}, \\ & \texttt{<|end-latent|>}, \texttt{<think>} appear \\
         & after the first latent block. \\
\addlinespace
$r = 0$ & Otherwise. \\
\bottomrule
\end{tabular}
\end{center}

\section{Latent RL Methodology}
\label{sec:appendix-latent-rl-methodology}

Latent RL method has two stages: 
\begin{enumerate}
    \item Training a value model $\rho_{\theta+\phi}$ by freezing parameters $\theta$.
    \item Training the policy model $\pi_\theta$ by leveraging the predictions of the value model for latent steps.
\end{enumerate}

In the first stage, data is collected to train $\rho_{\theta+\phi}$. A policy model with latent space thinking $\pi_\phi$ is used to generate responses to questions. For each response, an accuracy reward is computed using the reward function described in Section~\ref{sec:appendix-reward-funcs}. From these responses, the final-layer embeddings $\boldsymbol{h}$ are extracted and paired with the corresponding accuracy reward $r$, forming training pairs $(\boldsymbol{h}_i, r_i)$.

The value model $\rho_{\theta+\phi}$ maps the embedding $\boldsymbol{h}$ to an estimated reward $\rho_{\theta+\phi}(\boldsymbol{h})$. To train it, only the value head parameters $\phi$ are optimized by minimizing the binary cross-entropy (BCE) loss $L_{\text{BCE}}$ over embedding–reward pairs $(\boldsymbol{h}_i, r_i)$. Crucially, the loss is computed exclusively on embeddings corresponding to latent steps, while embeddings from other parts of the sequence—such as the question tokens or the final language answer—are ignored. This ensures that the value model learns to evaluate the quality of latent-space thinking steps.

\begin{equation}
    \arg\min\limits_{\phi} L_{BCE}(\rho_{\theta+\phi})
    \label{eq:value_head_bce_loss}
\end{equation}

\begin{equation}
\begin{split}
    L_{\text{BCE}}(\rho_{\theta+\phi}) 
    &= r_i \log \!\big(\rho_{\theta+\phi}(\boldsymbol{h}_i)\big) \\
    &\quad+\; \big(1 - r_i\big)\, \log \!\big(1 - \rho_{\theta+\phi}(\boldsymbol{h}_i)\big)
\end{split}
\end{equation}

In the second stage, the policy model $\pi_\theta$ is provided with a question and generates a sequence consisting of latent steps followed by a final language answer. This complete sequence—comprising the given question, the latent steps, and the generated answer—is then passed to the value model $\rho_{\theta+\phi}$, ensuring that it receives the same input as the policy model. Although the value model outputs estimated rewards for all positions in the sequence, the training objective is applied only to the latent step positions. The goal is to maximize the estimated reward of the latent steps, which is equivalent to minimizing $L_{LAT}$ in Equiation~\eqref{eq:latent-rl-objective}.

When optimizing Equation~\eqref{eq:latent-rl-objective}, the value head parameters $\phi$ are frozen. This is necessary because if $\phi$ were also updated, the value head could trivially learn to always predict high rewards, thereby collapsing the training signal. Instead, only the shared parameters $\theta$ are updated, which means that optimizing the value model $\rho_{\theta+\phi}$ also updates the policy model $\pi_\theta$, since the two models share $\theta$.

After this update step, the value head parameters $\phi$ are refined separately using the BCE objective in Equation~\eqref{eq:value_head_bce_loss}, with the true accuracy reward from that step as the target. This ensures that the value head remains aligned with the updated policy and continues to provide reliable reward estimates.

\section{Training Hyperparameters}
\label{sec:appendix-hps}

The training set ups are presented in Table~\ref{tab:sft_training_hps} and ~\ref{tab:grpo_training_hps}. All training runs use AdamW optimizer \citep{loshchilov2017decoupled}.

\begin{table*}[htbp]
    \centering
    \small
    \begin{tabular}{@{}l|l|l@{}}
    \toprule
    \textbf{Hyperparameter} & \textbf{Qwen SFT} & \textbf{LaST SFT} \\
    \midrule
    \textbf{Dataset} & GSM8K-Aug-NL & GSM8K-Aug-NL  \\
    \textbf{Num Examples} & 300K & 300K  \\
    \textbf{Base Model} & Qwen Base & Qwen Base \\
    \midrule
    \textbf{Hardware} & 4 × Nvidia A100-80GB & 8 × Nvidia A100-40GB \\
    \textbf{Training Time} & $\sim$ 1 hour & $\sim$ 8 hours \\
    \midrule
    \textbf{Num Epochs} & 1 & 5 \\
    \textbf{Per Device Batch Size} & 4 & 1 \\
    \textbf{Gradient Accum. Steps} & 32 & 128 \\
    \textbf{Effective Batch Size} & 512 & 512 \\
    \midrule
    \textbf{Learning Rate} & 5e-5 & 5e-5 \\
    \textbf{LR Scheduler} & StepLR & StepLR \\
    \textbf{LR Scheduler Args} & $\gamma=0.8$, step size=50 & $\gamma=0.8$, step size=50 \\
    \textbf{Warmup Steps} & -- & -- \\
    \textbf{Weight Decay} & 0.01 & 0.01 \\
    \bottomrule
    \end{tabular}
\caption{Training hyperparameters for Qwen SFT and LaST SFT models. Note that the optimizer and scheduler reset after each stage in LaST SFT training.}
\label{tab:sft_training_hps}
\end{table*}

\begin{table*}[htbp]
    \centering
    \small
    \setlength{\tabcolsep}{2pt}
    \begin{tabular}{@{}l|l|l|l@{}}
    \toprule
    \textbf{Hyperparameter} & \textbf{Qwen SFT \& GRPO} & \textbf{LaST SFT \& GRPO} & \textbf{LaST SFT \& Latent RL} \\
    \midrule
    \textbf{Dataset} & OpenR1-Math-220k & OpenR1-Math-220k & OpenR1-Math-220k \\
    \textbf{Num Examples} & 10K & 10K  & 10K  \\
    \textbf{Base Model} & Qwen SFT & LaST SFT & LaST SFT\\
    \midrule
    \textbf{Hardware} & 4 × Nvidia A100-80GB & 4 × Nvidia A100-80GB & 1 × Nvidia A100-40GB \\
    \textbf{Training Time} & $\sim$ 4 hours & $\sim$ 1 hour & $\sim$ 1.5 hours \\
    \midrule
    \textbf{Num Epochs} & 1 & 1 & 1 \\
    \textbf{Batch Size / Device} & 2 & 4 & 2\\
    \textbf{Grad. Accum. Steps} & 32 & 16 & 128 \\
    \textbf{Effective Batch Size} & 256 & 256 & 256 \\
    \midrule
    \textbf{Learning Rate} & 5e-6 & 5e-6 & 5e-6 \\
    \textbf{LR Scheduler} & Cosine With Min LR & Cosine With Min LR & Cosine With Min LR \\
    \textbf{LR Scheduler Args} & min lr rate = 0.1 & min lr rate = 0.1 & min lr rate = 0.1 \\
    \textbf{Warmup Steps} & 10 & 10 & 10\\
    \textbf{Weight Decay} & 0.01 & 0.01 & 0.01\\
    \midrule
    \textbf{Max Completion Len.} & 1024 & 1024 & 1024\\
    \textbf{Temperature} & 0.7 & 0.7 & 0\\
    \textbf{Num Generations} & 8 & 8 & 1 \\
    \textbf{KL Div. Coeff. $\beta$} & 0.0 & 0.64 & - \\
    \bottomrule
    \end{tabular}
\caption{Training hyperparameters for Qwen SFT \& GRPO and LaST SFT \& GRPO models. Note that the effective batch size contains $256$ responses to $256/8=32$ different questions and $8$ responses for each questions in GRPO training.}
\label{tab:grpo_training_hps}
\end{table*}

\section{Instruction Following}
\label{sec:appendix-instruction-following}

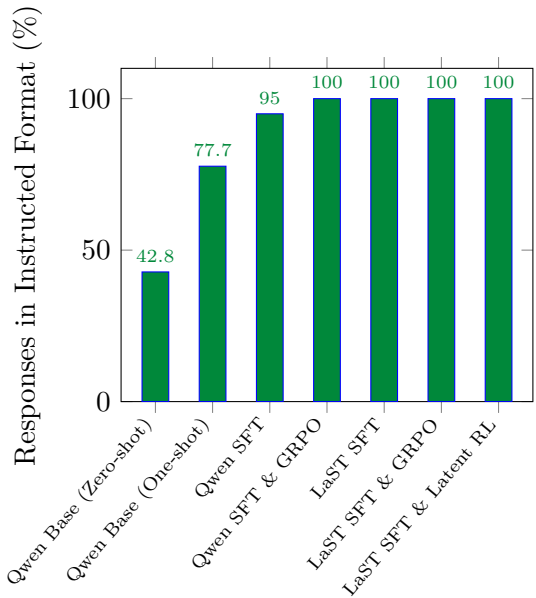
\begin{figure}[ht]
\centering
\begin{tikzpicture}
    \begin{axis}[
        ybar,
        bar width=10pt,
        width=7cm,
        height=6cm,
        ymin=0, ymax=110,
        ylabel={Responses in Instructed Format (\%)},
        symbolic x coords={
            Qwen Base (Zero-shot),
            Qwen Base (One-shot),
            Qwen SFT,
            Qwen SFT \& GRPO,
            LaST SFT,
            LaST SFT \& GRPO,
            LaST SFT \& Latent RL
        },
        xtick=data,
        x tick label style={
            rotate=50,
            font=\scriptsize,
            anchor=east
        },
        nodes near coords,
        nodes near coords align={vertical},
        every node near coord/.append style={font=\scriptsize, text=LMUGreen},
        ymin=0
    ]
        \addplot+[ybar, fill=LMUGreen] coordinates {
            (Qwen Base (Zero-shot),42.8)
            (Qwen Base (One-shot),77.7)
            (Qwen SFT,95)
            (Qwen SFT \& GRPO,100)
            (LaST SFT,100)
            (LaST SFT \& GRPO,100)
            (LaST SFT \& Latent RL,100)
        };
    \end{axis}
\end{tikzpicture}
\caption{Ratio of answers following the instructed format during evaluation of models in GSM8K and MATH500 benchmarks. All models evaluated in zero-shot setting unless otherwise indicated.}
\label{fig:ratio_of_instruction_following}    
\end{figure}

The ratio of generated responses that are fitting to the instructed format is given in Figure~\ref{fig:ratio_of_instruction_following}.

\section{Latent RL Experiments}
\label{sec:appendix-latent-rl-training-curves}

The training curves in Latent RL training procedure for the LaST SFT \& Latent RL model in in Table~\ref{tab:overall_performance_results} are presented in this section. 

\begin{figure*}[ht]
    \centering
    \subcaptionbox{%
        \label{fig:latent_rl_first_phase_training_loss}
    }[0.49\textwidth]{%
        \includegraphics[width=\linewidth]{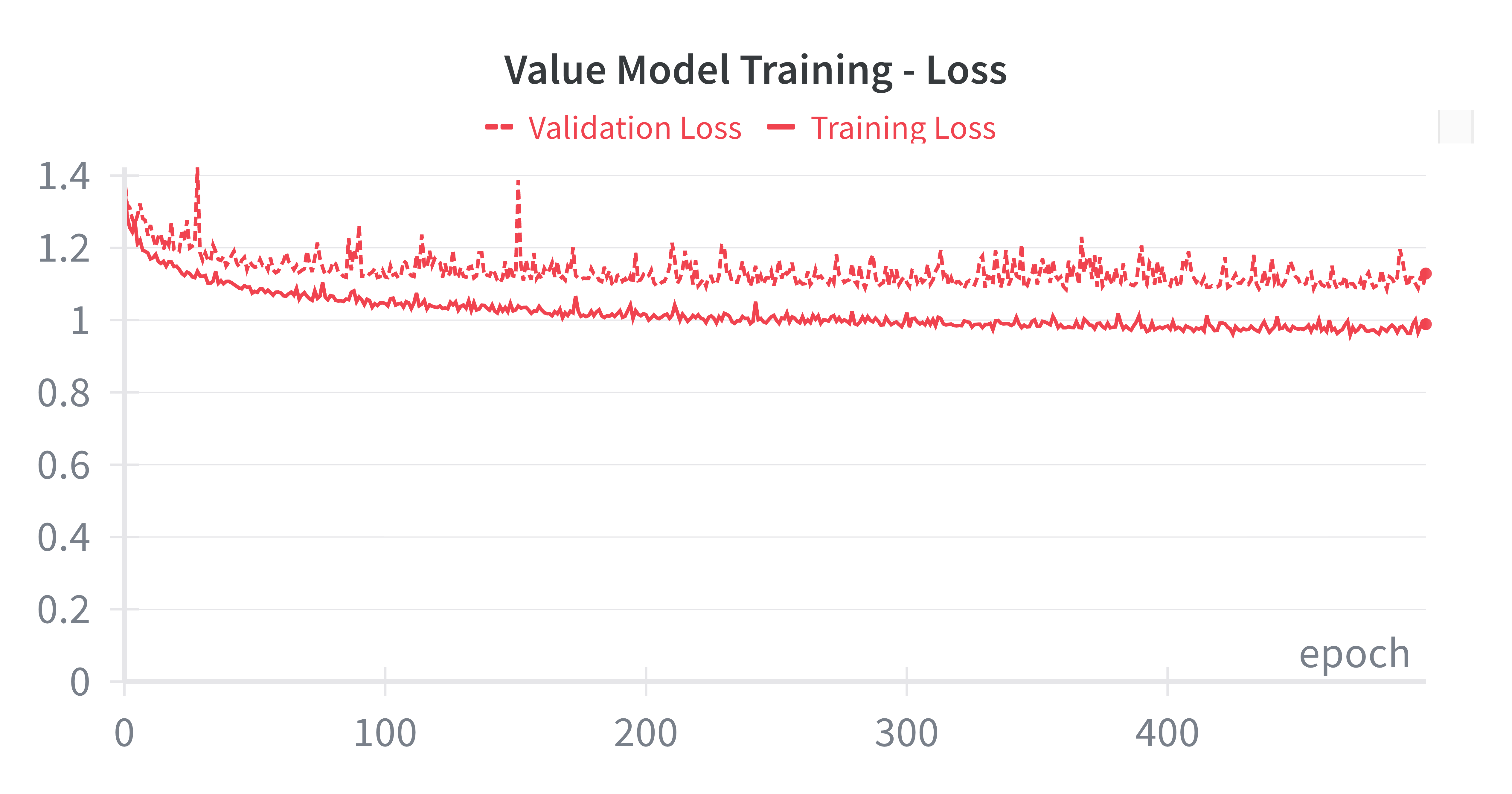}
    }%
    \hspace{0.01\textwidth}
    \subcaptionbox{%
        \label{fig:latent_rl_first_phase_training_f1_rocauc}
    }[0.49\textwidth]{%
        \includegraphics[width=\linewidth]{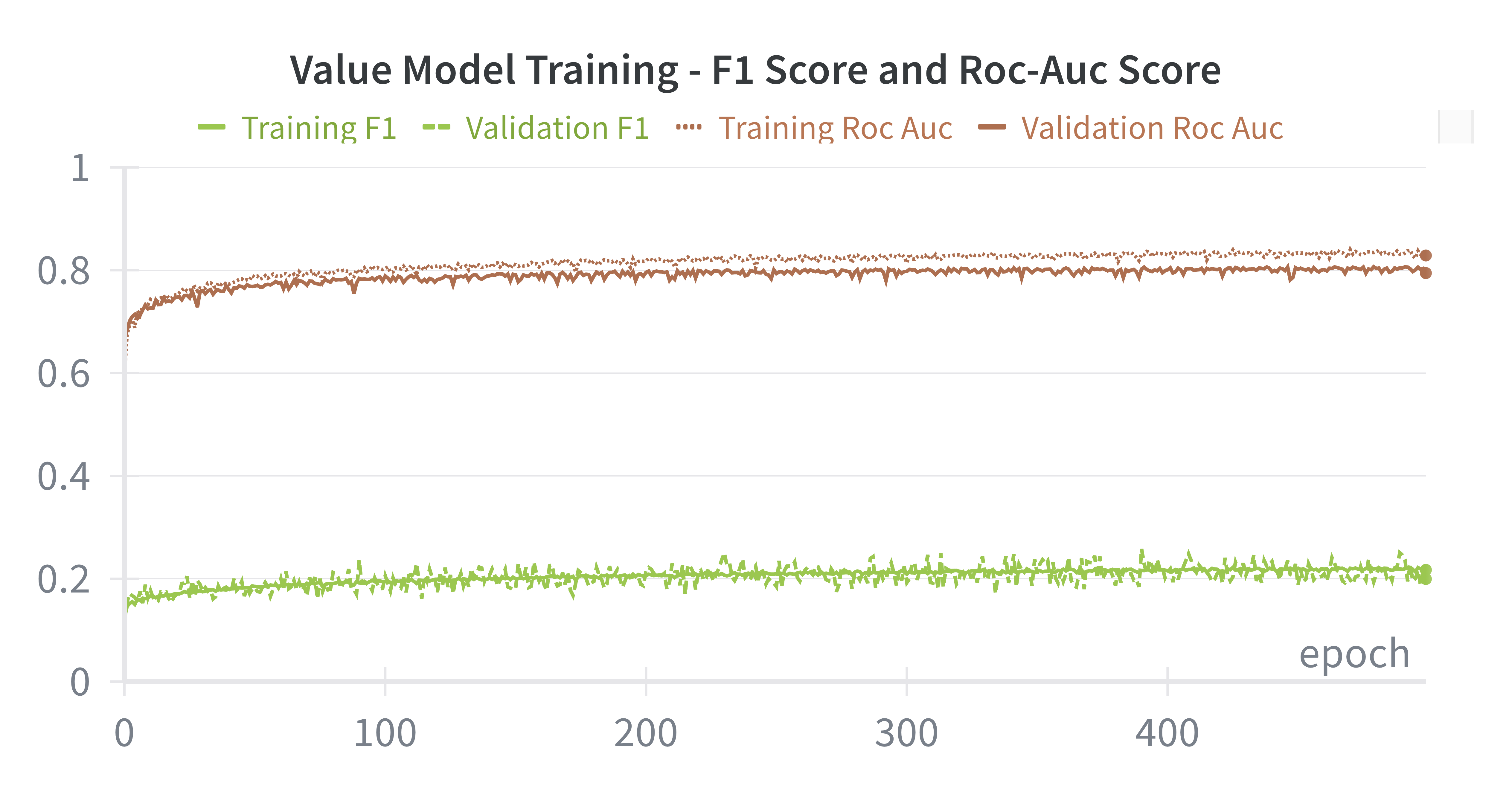}
    }
    \caption{Training curves for the value model training in the first stage of Latent RL. Note that only the linear value head is trained. 
    (a) BCE loss.
    (b) F1 and Roc Auc scores.
    }
    \label{fig:latent_rl_first_phase_training}
\end{figure*}

As described in Section~\ref{sec:last-latentrl}, in the first phase we train a value model (Figure~\ref{fig:latent_rl_first_phase_training}). Although the model achieves a high ROC-AUC score, its F1 score remains comparatively lower. This discrepancy arises because ROC-AUC measures the overall ranking ability of the model across thresholds, whereas the F1 score reflects the balance of precision and recall at a fixed threshold. Since the second phase of Latent RL training relies on the predicted probabilities rather than thresholded labels, ROC-AUC is a more informative indicator of value head performance in this setting.

\begin{figure*}[ht]
    \centering
    \subcaptionbox{%
        \label{fig:latent_rl_policy_loss}
    }[0.49\textwidth][c]{%
        \includegraphics[width=\linewidth]{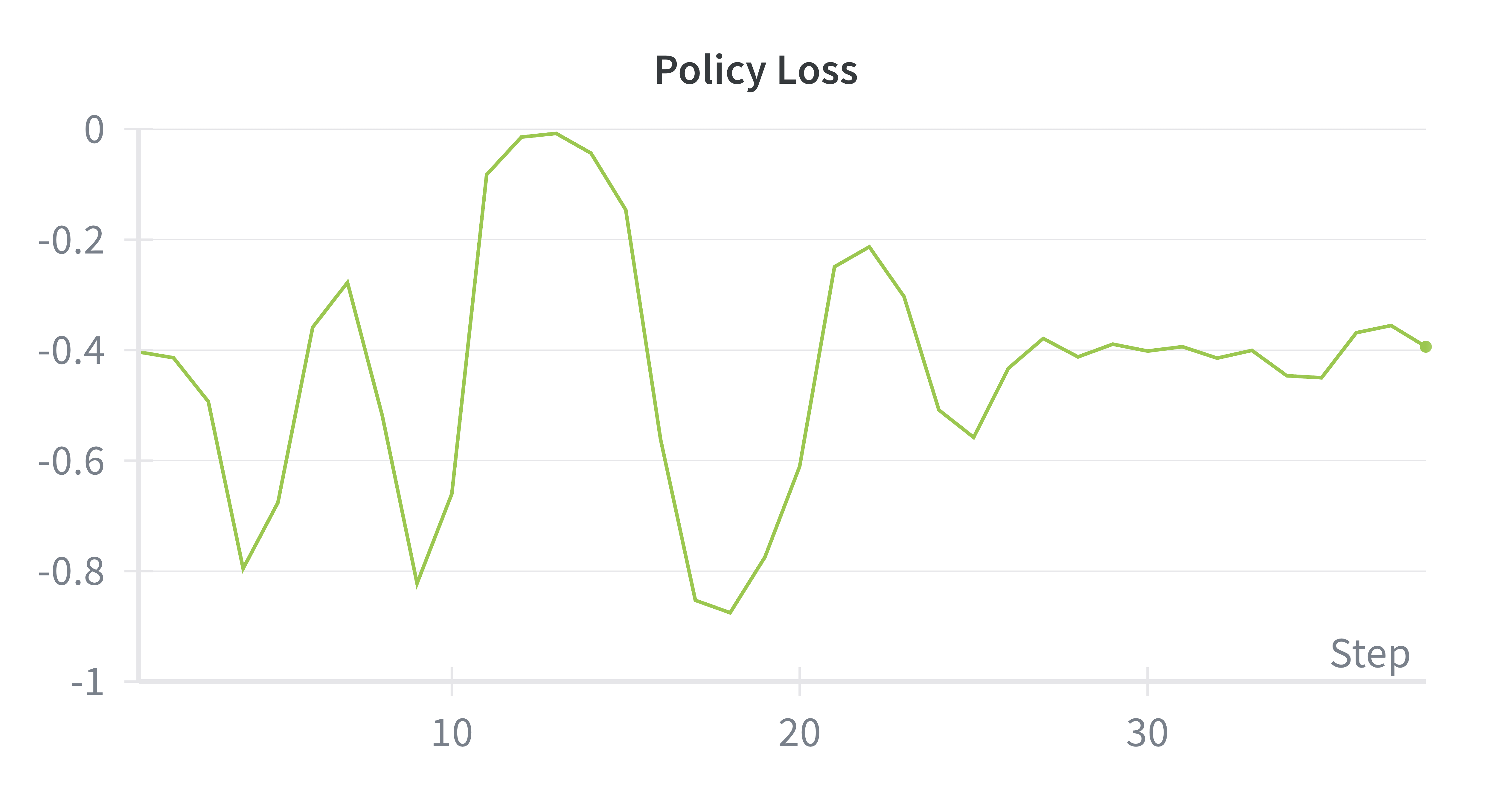}
    }%
    \hspace{0.01\textwidth}
    \subcaptionbox{%
        \label{fig:latent_rl_value_head_loss}
    }[0.49\textwidth][c]{%
        \includegraphics[width=\linewidth]{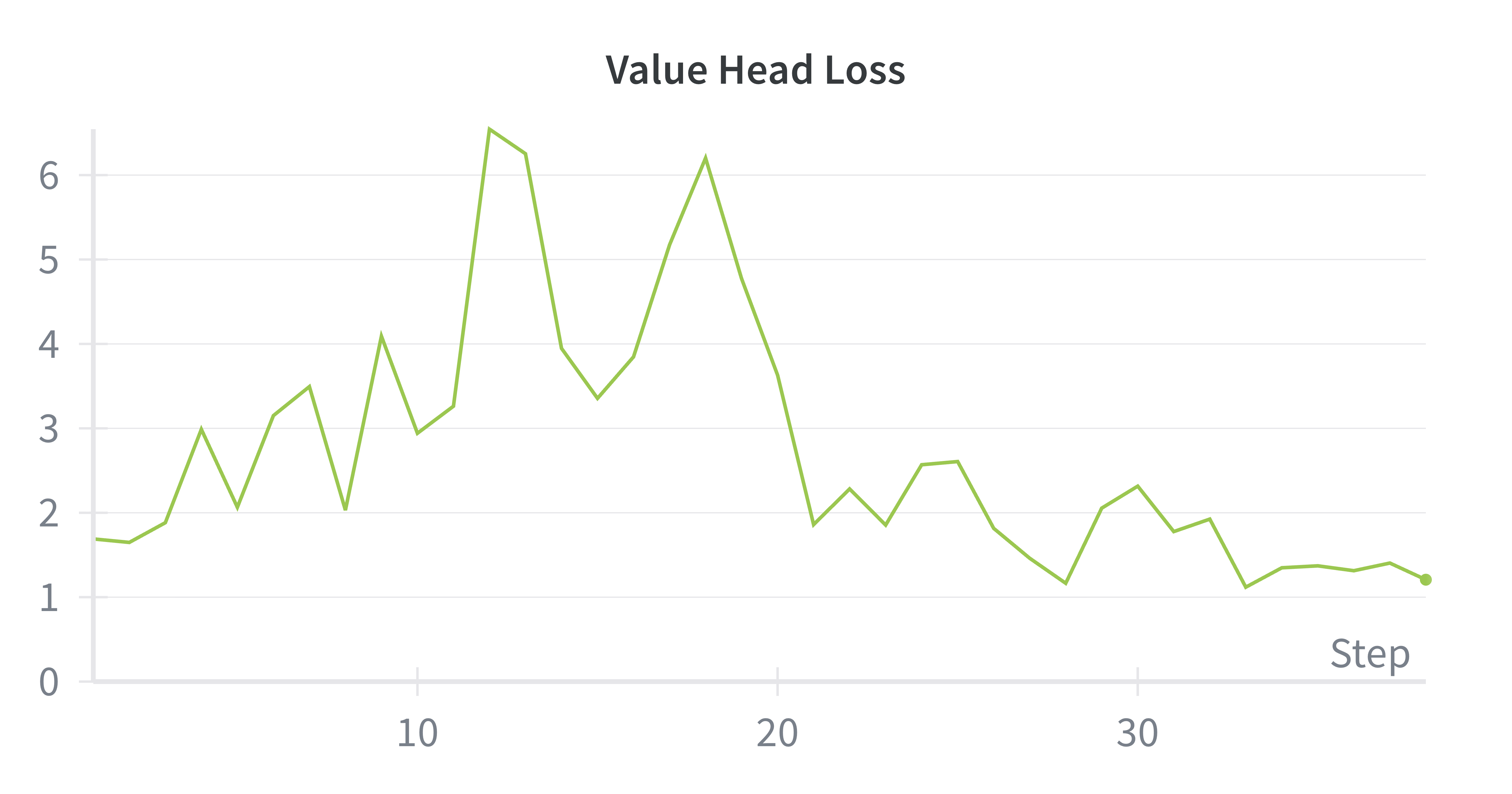}
    }%
    \vspace{0.01em}
    \subcaptionbox{%
        \label{fig:latent_rl_rewards}
    }[0.49\textwidth][c]{%
        \includegraphics[width=\linewidth]{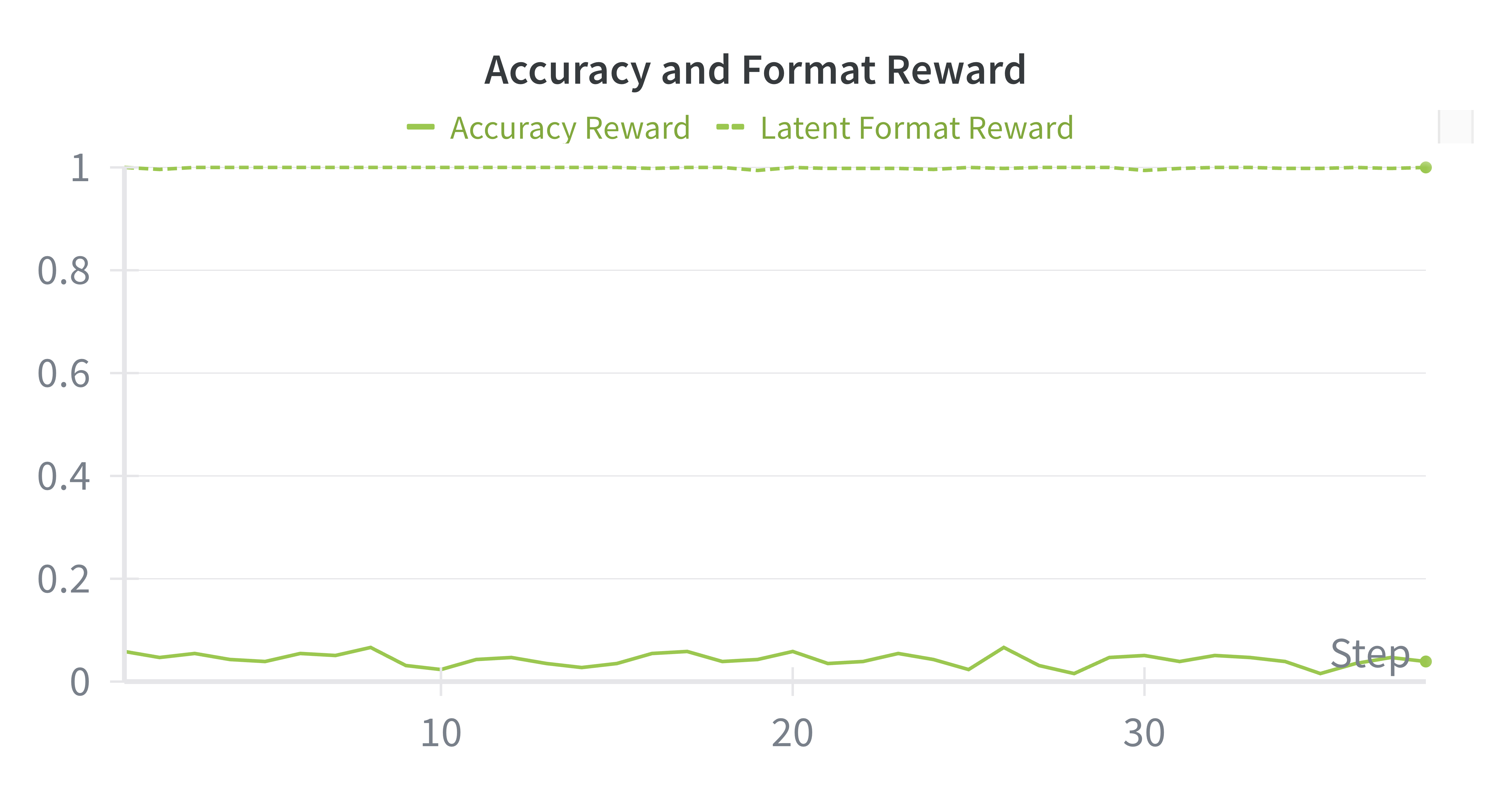}
    }
\caption{Training details of Latent RL: (a) policy loss, (b) value head BCE loss, (c) accuracy and latent format rewards.}
    \label{fig:latent_rl_training_curves}
\end{figure*}

After training the value model, in the second stage both the policy model and the value models are trained sequentially. However, we see that the both policy loss and value head loss are fluctuating, indicating an unstable training (Figure~\ref{fig:latent_rl_training_curves}).

\begin{figure*}[ht]
    \centering
    \subcaptionbox{%
        \label{fig:latent_rl_frozen_value_model_policy_loss}
    }[0.49\textwidth]{%
        \includegraphics[width=\linewidth]{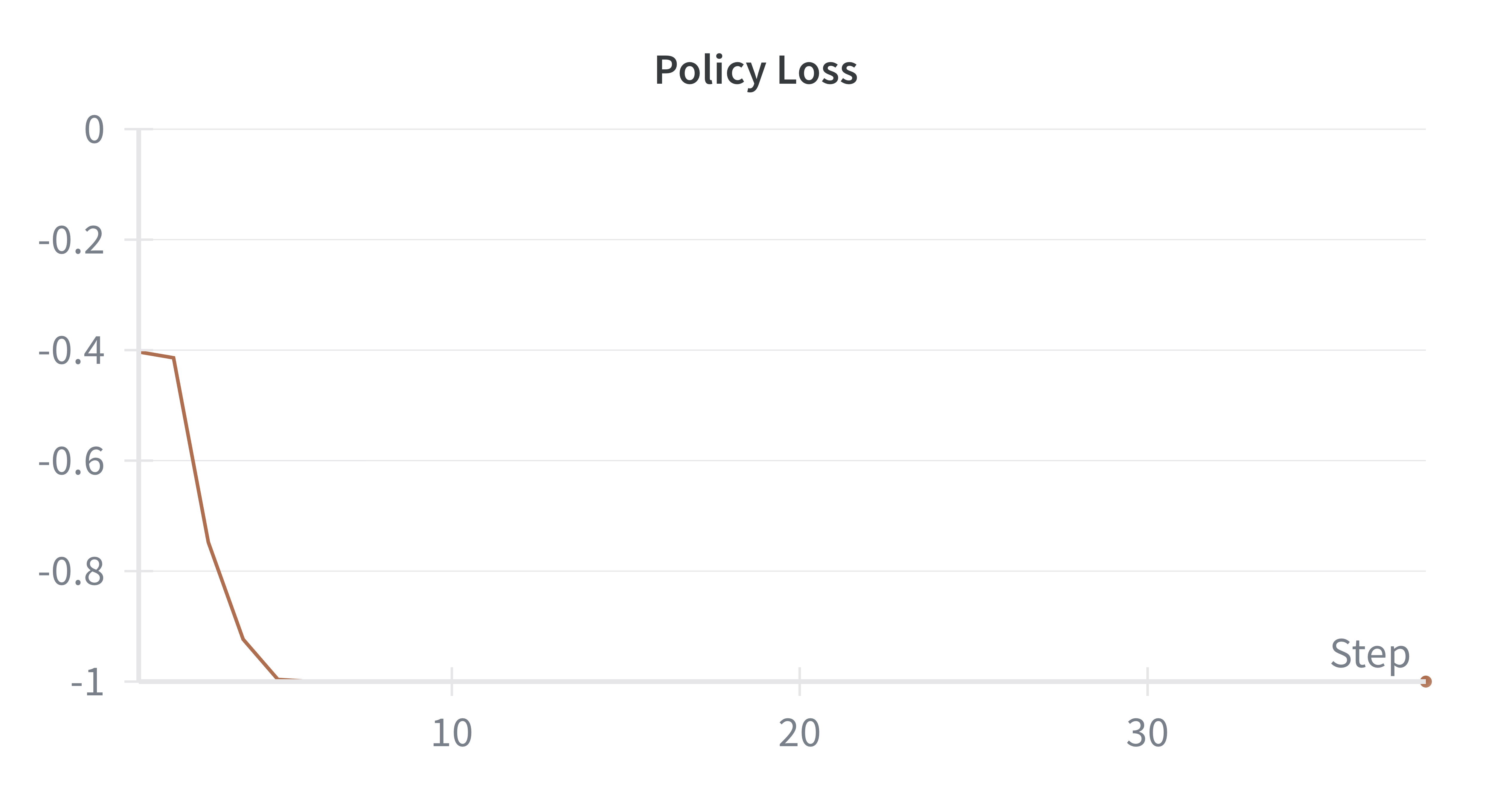}
    }%
    \hspace{0.01\textwidth}
    \subcaptionbox{%
        \label{fig:latent_rl_frozen_value_model_value_model_loss}
    }[0.49\textwidth]{%
        \includegraphics[width=\linewidth]{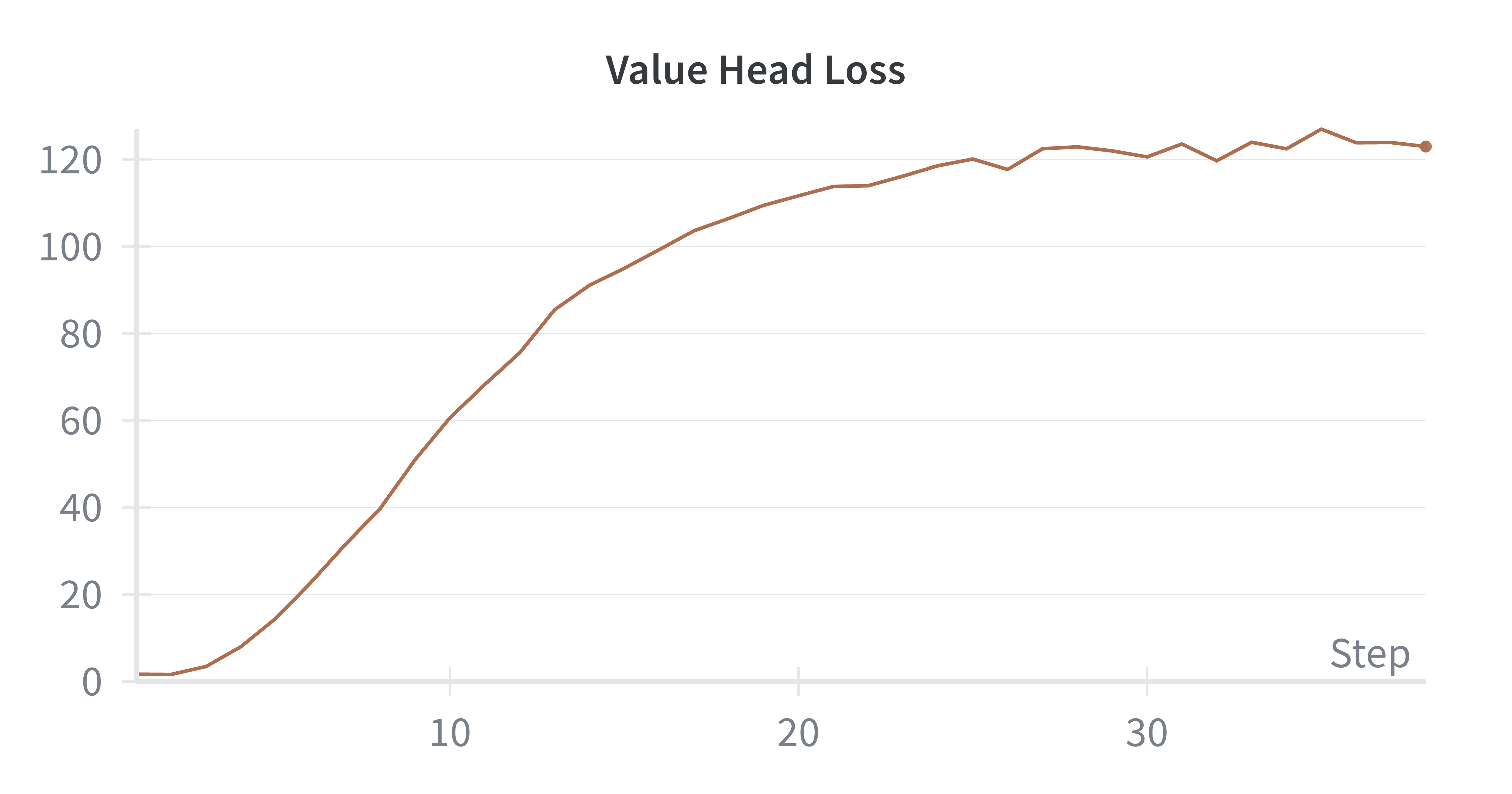}
    }
    \subcaptionbox{%
        \label{fig:latent_rl_frozen_value_model_rewards}
    }[0.49\textwidth]{%
        \includegraphics[width=\linewidth]{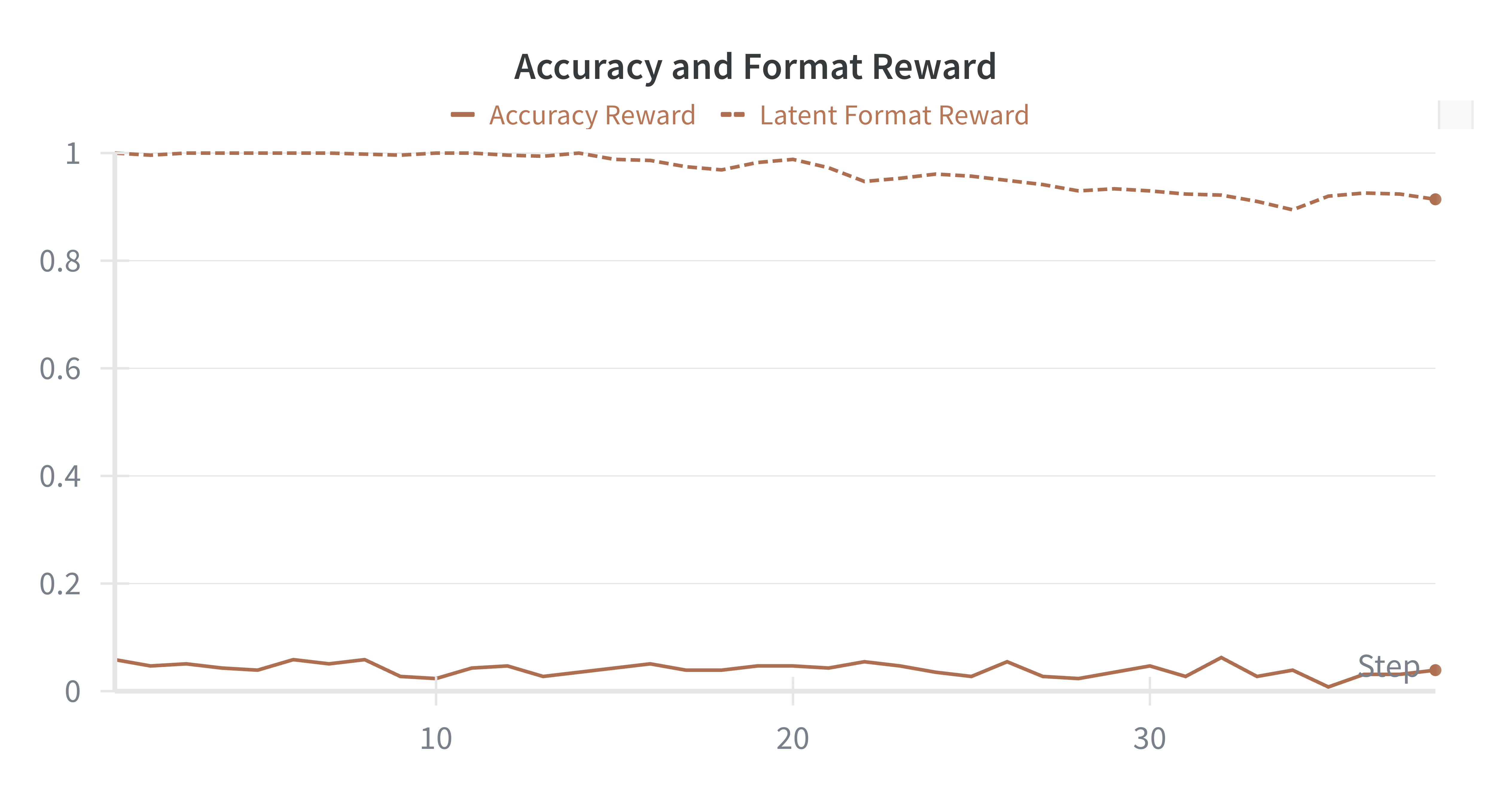}
    }%
    \caption{Training dynamics of Latent RL with a frozen value model. 
    (a) Policy loss shows smooth convergence. 
    (b) Value model loss, monitored but not optimized. 
    (c) Accuracy and latent format rewards.
    }
    \label{fig:latent_rl_frozen_value_model_training_curves}
\end{figure*}

To stabilize training, we freeze the value head during the second phase of Latent RL. This prevents it from becoming a moving target and leads to smoother optimization for the policy model, as shown in Figure~\ref{fig:latent_rl_frozen_value_model_training_curves}. Although the policy loss decreases steadily, overall accuracy does not improve.

This stagnation arises from two causes: misalignment between the frozen value head and the evolving policy, and reward hacking. As the policy changes, the static value head becomes less accurate, while the policy exploits its fixed reward signal instead of improving reasoning or task accuracy. Thus, smoother training does not yield better benchmark performance.

\section{Example Responses from MATH500 Benchmark}
\label{sec:appendix-ex-responses}

Example answers from the models reported in Table~\ref{tab:overall_performance_results} are shown in Figure~\ref{fig:model_ex_responses_qwen_base} to Figure~\ref{fig:model_ex_responses_last_sft_latenrl} for a representative question from the MATH500 benchmark, illustrating their behavior.

\begin{figure}[htbp]
    \centering
    \begin{minipage}{0.47\textwidth}
        \begin{tcolorbox}[
            colback=blue!10,
            colframe=blue!40,
            boxrule=0.8pt,
            arc=3pt,
            left=8pt,
            right=8pt,
            top=1pt,
            bottom=1pt,
            title={\textbf{Question}},
            fonttitle=\small\sffamily,
            coltitle=blue!80!black
        ]
        \begin{lstlisting}[
            breaklines=true,
            breakindent=0pt,
            aboveskip=0pt, belowskip=0pt,
            breakatwhitespace=true,
            basicstyle=\scriptsize\ttfamily,
            frame=none,
            xleftmargin=0pt,
            xrightmargin=0pt
        ]
How many positive whole-number divisors does 196 have?       
        \end{lstlisting}
        \end{tcolorbox}
    \end{minipage}
    \caption{Example question from MATH500 benchmark. The question, inserted into the system prompt which is omitted here for brevity. The correct answer is 9.}
    \label{fig:model_ex_responses_question}
\end{figure}

\begin{figure}[htbp]
    \centering
    \begin{minipage}{0.49\textwidth}
        \begin{tcolorbox}[
            colback=gray!5,
            colframe=gray!30,
            boxrule=0.8pt,
            arc=3pt,
            left=8pt,
            right=8pt,
            top=1pt,
            bottom=1pt,
            title={\textbf{Qwen Base Response}},
            fonttitle=\small\sffamily,
            coltitle=gray!70!black
        ]
        \begin{lstlisting}[
            breaklines=true,
            breakindent=0pt,
            aboveskip=0pt, belowskip=0pt,
            breakatwhitespace=true,
            basicstyle=\scriptsize\ttfamily,
            frame=none,
            xleftmargin=0pt,
            xrightmargin=0pt
        ]
<think>196 is a perfect square, so it has an even number of divisors. The prime factorization of 196 is 2^2 * 7^2. The number of divisors is (2+1)(2+1) = 9. So, 196 has 9 positive whole-number divisors.</think><answer>\boxed{9}</answer>
        \end{lstlisting}
        \end{tcolorbox}
    \end{minipage}
    \caption{Example Qwen Base (zero-shot) response to the question in Figure~\ref{fig:model_ex_responses_question}.}
    \label{fig:model_ex_responses_qwen_base}
\end{figure}

\begin{figure}[htbp]
    \centering
    \begin{minipage}{0.49\textwidth}
        \begin{tcolorbox}[
            colback=gray!5,
            colframe=gray!30,
            boxrule=0.8pt,
            arc=3pt,
            left=8pt,
            right=8pt,
            top=1pt,
            bottom=1pt,
            title={\textbf{Qwen SFT Response}},
            fonttitle=\small\sffamily,
            coltitle=gray!70!black
        ]
        \begin{lstlisting}[
            breaklines=true,
            breakindent=0pt,
            aboveskip=0pt, belowskip=0pt,
            breakatwhitespace=true,
            basicstyle=\scriptsize\ttfamily,
            frame=none,
            xleftmargin=0pt,
            xrightmargin=0pt
        ]
<think>First prime factorize 196=2^2*7^2. The prime factorization of any divisor of 196 can only include the primes 2 and 7. 
We are free to choose either 0, 1, 2, or 3 as the exponent of 2 (4 options).
For the exponent of 7, we can choose 0 or 1 (2 options).
In total, there are $4\times 2=8$ possibilities.</think><answer>\boxed{8}</answer>
        \end{lstlisting}
        \end{tcolorbox}
    \end{minipage}
    \caption{Example Qwen SFT response to the question in Figure~\ref{fig:model_ex_responses_question}.}
    \label{fig:model_ex_responses_qwen_sft}
\end{figure}

\begin{figure}[htbp]
    \centering
    \begin{minipage}{0.49\textwidth}        
        \begin{tcolorbox}[
            colback=gray!5,
            colframe=gray!30,
            boxrule=0.8pt,
            arc=3pt,
            left=8pt,
            right=8pt,
            top=1pt,
            bottom=1pt,
            title={\textbf{Qwen SFT \& GRPO Response}},
            fonttitle=\small\sffamily,
            coltitle=gray!70!black
        ]
        \begin{lstlisting}[
            breaklines=true,
            breakindent=0pt,
            aboveskip=0pt, belowskip=0pt,
            breakatwhitespace=true,
            basicstyle=\scriptsize\ttfamily,
            frame=none,
            xleftmargin=0pt,
            xrightmargin=0pt
        ]
<think>The problem requires us to find the number of positive whole-number divisors of 196. First, we need to determine the prime factorization of 196. We start by dividing 196 by the smallest prime number, which is 2:

\[196 \div 2 = 98\]

Next, we divide 98 by 2:

\[98 \div 2 = 49\]

Now, 49 is not divisible by 2, so we check the next smallest prime number, which is 7:

\[49 \div 7 = 7\]

And again, 7 is a prime number. Therefore, the prime factorization of 196 is:

\[196 = 2^2 \times 7^2\]

To find the number of positive whole-number divisors, we use the formula that if a number has a prime factorization of the form \(p_1^{e_1} \times p_2^{e_2} \times \cdots \times p_k^{e_k}\), then the number of divisors is given by \((e_1 + 1)(e_2 + 1) \cdots (e_k + 1)\). For the factorization \(2^2 \times 7^2\), the exponents are 2 and 2. Adding 1 to each of these exponents gives 3 and 3. Therefore, the number of divisors is:

\[(2+1)(2+1) = 3 \times 3 = 9\]

Thus, the number of positive whole-number divisors of 196 is \(\boxed{9}\).</think><answer>\boxed{9}</answer>
        \end{lstlisting}
        \end{tcolorbox}        
    \end{minipage}
    \caption{Example Qwen SFT \& GRPO response to the question in Figure~\ref{fig:model_ex_responses_question}.}
    \label{fig:model_ex_responses_qwen_sft_grpo}
\end{figure}

\begin{figure}[htbp]
    \centering
    \begin{minipage}{0.49\textwidth}        
        \begin{tcolorbox}[
            colback=gray!5,
            colframe=gray!30,
            boxrule=0.8pt,
            arc=3pt,
            left=8pt,
            right=8pt,
            top=1pt,
            bottom=1pt,
            title={\textbf{LaST SFT Response}},
            fonttitle=\small\sffamily,
            coltitle=gray!70!black
        ]
        \begin{lstlisting}[
            breaklines=true,
            breakindent=0pt,
            aboveskip=0pt, belowskip=0pt,
            breakatwhitespace=true,
            basicstyle=\scriptsize\ttfamily,
            frame=none,
            xleftmargin=0pt,
            xrightmargin=0pt
        ]
<|start-latent|><|latent|><|latent|><|latent|>
<|latent|><|latent|><|latent|><|end-latent|>
<answer>\boxed{14}</answer>
        \end{lstlisting}
        \end{tcolorbox}
    \end{minipage}
    \caption{Example LaST SFT response to the question in Figure~\ref{fig:model_ex_responses_question}.}
    \label{fig:model_ex_responses_last_sft}
\end{figure}

\begin{figure}[htbp]
    \centering
    \begin{minipage}{0.49\textwidth}
        \begin{tcolorbox}[
            colback=gray!5,
            colframe=gray!30,
            boxrule=0.8pt,
            arc=3pt,
            left=8pt,
            right=8pt,
            top=1pt,
            bottom=1pt,
            title={\textbf{LaST SFT \& GRPO Response}},
            fonttitle=\small\sffamily,
            coltitle=gray!70!black
        ]
        \begin{lstlisting}[
            breaklines=true,
            breakindent=0pt,
            aboveskip=0pt, belowskip=0pt,
            breakatwhitespace=true,
            basicstyle=\scriptsize\ttfamily,
            frame=none,
            xleftmargin=0pt,
            xrightmargin=0pt
        ]
<|start-latent|><|latent|><|latent|><|latent|>
<|latent|><|latent|><|latent|><|end-latent|>
<answer>\boxed{14}</answer>
        \end{lstlisting}
        \end{tcolorbox}
    \end{minipage}
    \caption{Example LaST SFT \& GRPO response to the question in Figure~\ref{fig:model_ex_responses_question}.}
    \label{fig:model_ex_responses_qwen_last_sft_grpo}
\end{figure}

\begin{figure}[htbp!]
    \centering
    \begin{minipage}{0.49\textwidth}
        \begin{tcolorbox}[
            colback=gray!5,
            colframe=gray!30,
            boxrule=0.8pt,
            arc=3pt,
            left=8pt,
            right=8pt,
            top=1pt,
            bottom=1pt,
            title={\textbf{LaST SFT \& Latent RL Response}},
            fonttitle=\small\sffamily,
            coltitle=gray!70!black
        ]
        \begin{lstlisting}[
            breaklines=true,
            breakindent=0pt,
            aboveskip=0pt, belowskip=0pt,
            breakatwhitespace=true,
            basicstyle=\scriptsize\ttfamily,
            frame=none,
            xleftmargin=0pt,
            xrightmargin=0pt
        ]
<|start-latent|><|latent|><|latent|><|latent|>
<|latent|><|latent|><|latent|><|end-latent|>
<answer>\boxed{14}</answer>
        \end{lstlisting}
        \end{tcolorbox}
    \end{minipage}
    \caption{Example LaST SFT \& Latent RL response to the question in Figure~\ref{fig:model_ex_responses_question}.}
    \label{fig:model_ex_responses_last_sft_latenrl}
\end{figure}

For this MATH500 example, the language-space models show variation in their responses, whereas all latent-space thinking models produce identical final answer. Since the latent steps are continuous, changes in their behavior cannot easily be inferred without additional interpretability studies.

\end{document}